\pdfoutput=1

\documentclass[11pt]{article}

\usepackage[final]{acl}

\usepackage{times}
\usepackage{latexsym}
\usepackage{amsmath, amsfonts}

\usepackage[T1]{fontenc}

\usepackage[utf8]{inputenc}

\usepackage{microtype}

\usepackage{inconsolata}

\usepackage{graphicx}

\title{Does Self-Attention Need Separate Weights in Transformers?}

\author{
\textbf{Md Kowsher}\textsuperscript{1}, 
\textbf{Nusrat Jahan Prottasha}\textsuperscript{1}, 
\textbf{Chun-Nam Yu} \textsuperscript{2} \\
\textbf{Ozlem Ozmen Garibay}\textsuperscript{1}, 
\textbf{Niloofar Yousefi}\textsuperscript{1} \\
\textsuperscript{1}University of Central Florida, FL, USA \\
\textsuperscript{2}Nokia Bell Labs, NJ, USA \\
 \\
}

\begin{document}
\maketitle

\begin{abstract}
Self-attention has revolutionized natural language processing by capturing long-range dependencies and improving context understanding. However, it comes with high computational costs and struggles with sequential data's inherent directionality. This paper investigates and presents a simplified approach called "shared weight self-attention," where a single weight matrix is used for Keys, Queries, and Values instead of separate matrices for each. This approach cuts training parameters by more than half and significantly reduces training time. Our method not only improves efficiency but also achieves strong performance on tasks from the GLUE benchmark, even outperforming the standard BERT baseline in handling noisy and out-of-domain data. Experimental results show a 66.53\% reduction in parameter size within the attention block and competitive accuracy improvements of 3.55\% and 0.89\% over symmetric and pairwise attention-based BERT models, respectively. 
\end{abstract}

\section{Introduction}

Natural language processing (NLP) has seen remarkable progress with the advent of transformer-based architectures \cite{gillioz2020overview, kowsher2022bangla}. These models have revolutionized tasks such as machine translation \cite{lopez2008statistical}, language modeling \cite{jozefowicz2016exploring}, and question answering \cite{allam2012question, kowsher2024token}, achieving better accuracy and performance. Central to the success of these models is the self-attention mechanism \cite{vaswani2017attention, shaw2018self}, which allows them to weigh the importance of different words in a sentence dynamically.


Self-Attention's main challenges include computational inefficiency with quadratic complexity, difficulty in handling long-term dependencies effectively, and the lack of inherent directionality in capturing sequential relationships.
While the attention mechanism itself has been extensively investigated~\cite{bielik2020adversarial, choromanski2020rethinking, zhuang2023survey, phan2021cotext}, and improvements in computational complexity have been proposed \cite{kitaev2020reformer, zhu2020deformable, xiao2022retromae}, the primary method retains the same architecture in using separate trainable weight matrices to compute Keys, Queries, and Values, which leads to a high parameter count and significant complexity for computing attention.
We would like to ask: \textit{``Do we need the three weight matrix representations of (Key, Query, Value) for learning self-attention scores?" }

To address this question, we revisit the concept of self-attention and propose a novel shared weight self-attention mechanism that employs a single weight matrix for all three representations to reduce the parameter size and the time and memory complexity.
Our shared weight matrix enables the model to efficiently capture the essential features needed for understanding semantics without the overhead of managing multiple matrices. The shared matrix is a regularization to capture the common weights learned from each representation.  This simplification reduces the model's computational footprint, retains the ability to focus on relevant parts of the input data effectively, and enhances prediction generalization for noisy input and out-of-domain test data.

In this work, we explore alternative compatibility functions within the self-attention mechanism of Transformer-based encoder models, particularly BERT \cite{devlin2018BERT}.  By utilizing a shared representation for (Key, Query, Value), our approach achieves substantial improvements in efficiency while maintaining the model's performance without any compromise on accuracy. 

Our contributions can be summarized as follows: 

\begin{itemize}
    \item We introduce a new shared self-attention mechanism that employs a single weight matrix, $W_s$ for (Key, Query, Value). 

    \item Shared weight shows a 66.53\% reduction in self-attention block parameters and 12.94\% reduction in total BERT model parameters while maintaining performance across various downstream tasks.

\end{itemize}

\section{Shared Self-Attention}

\begin{figure*}[!t]

 \begin{center}
    \includegraphics[width=1.00\linewidth]{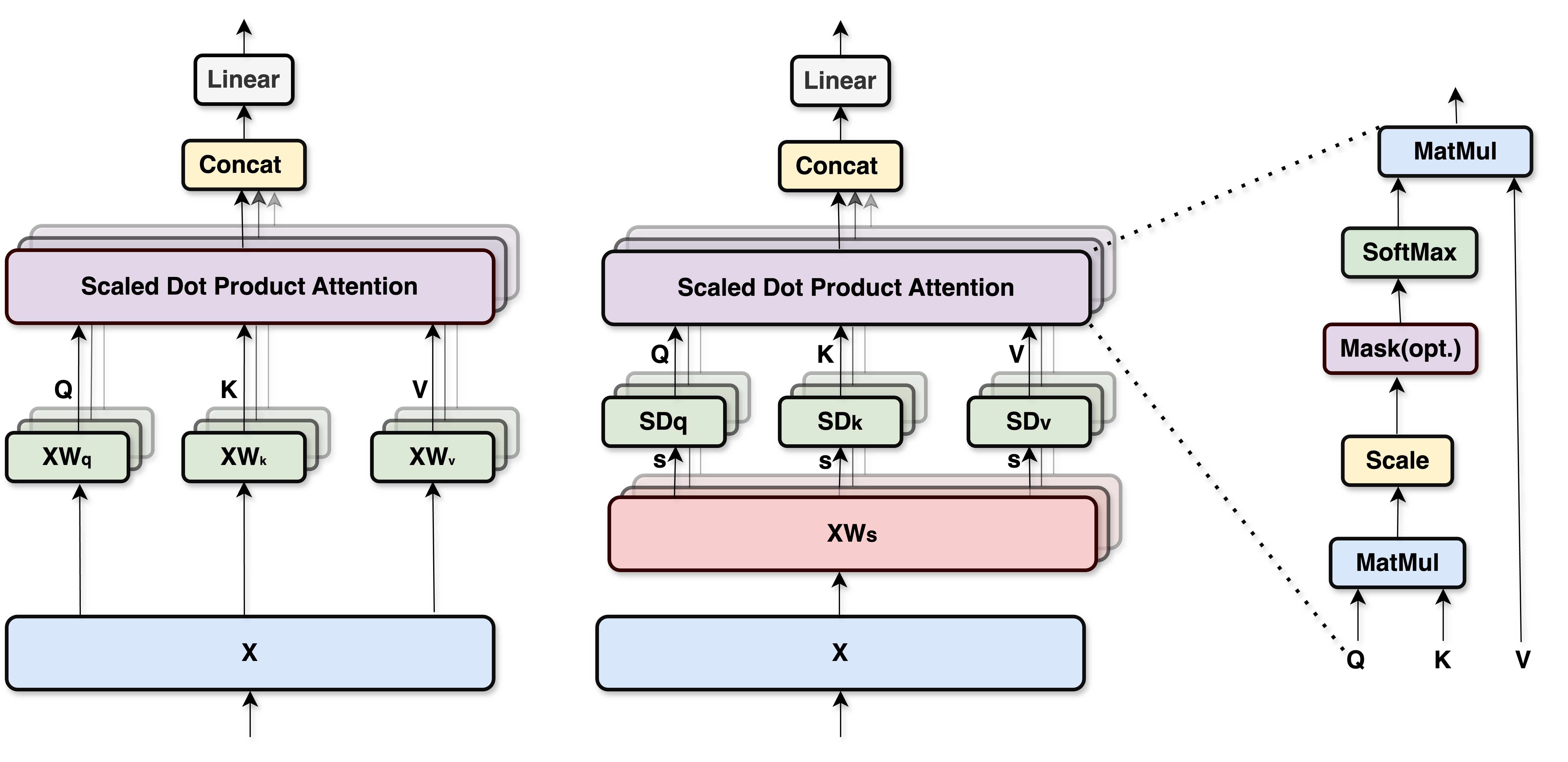}

   \end{center}
\caption{Comparison of traditional self-attention (left) and shared weight self-attention (right). }
\end{figure*}

\subsection{Preliminaries}

Consider an input matrix \( X \in \mathbb{R}^{n \times d} \), where \( n \) is the sequence length and \( d \) is the dimensionality of the input space. The self-attention mechanism traditionally maps this input into three distinct representations: keys \( K \), queries \( Q \), and values \( V \), using separate linear transformations with weight matrices \( W_k \), \( W_q \), and \( W_v \) respectively. We propose a unified representation using a single matrix \( W_s \) from which these mappings are derived, leading to a reduction in the number of parameters and accelerating the self-attention layer.

\subsection{Self-Attention}
In traditional self-attention, distinct linear transformations are employed to generate keys \( K \), queries \( Q \), and values \( V \) from the input \( X \). This process can be mathematically expressed as:
\[
K = X W_k, \quad Q = X W_q, \quad V = X W_v, 
\]
where \( W_k, W_q, W_v \in \mathbb{R}^{d \times d} \) are learnable weight matrices corresponding to keys, queries, and values respectively. These matrices allow the model to adaptively focus on different parts of the input by calculating attention weights through the softmax-normalized dot product of queries and keys:
\begin{equation}
\label{eq:self_att}
\text{Attention}(Q, K, V) = \text{softmax}\left(\frac{Q K^T}{\sqrt{d}}\right) V, 
\end{equation}
where \( d \) is the dimension of the model, which aids in stabilizing the learning process.

\subsection{Shared Weight Self-Attention}

We define a shared transformation function \( \mathbf{S}: \mathbb{R}^d \rightarrow \mathbb{R}^d \) parameterized by a weight matrix \( W_s \) containing learnable parameters:
\[
S = \mathbf{S}(X) = X W_s, \quad W_s \in \mathbb{R}^{d \times d}
\]
This function \(  \mathbf{S} \) is designed to capture the core semantic features of the input in a single compact representation \( S \).

To derive the keys, queries, and values vectors from the unified representation \( S \), we introduce three separate diagonal transformation matrices \( D_k, D_q, D_v \), each in \( \mathbb{R}^{d\times d} \). These diagonal matrices act as element-wise scaling factors that adapt the shared representation \( S \) for specific roles in the attention mechanism:
\begin{align*}
  Q &= S D_q = X W_s D_q \\
  K &= S D_k = X W_s D_k \\
  V &= S D_v = X W_s D_v
\end{align*}
This can be interpreted as having a special factorization of the weight matrices $W_q, W_k, W_v$ used in standard attention as $W_q = W_s D_q$, $W_k = W_s D_k$, and $W_v = W_s D_v$, where $W_s$ is shared and the diagonal $D_q, D_k, D_v$ reduce the parameter count and allow for efficient and differentiated modulation of the base representation \( S \). Now, we can calculate the attention score by following Equation \ref{eq:self_att}.  

\subsection{Experiments}

To evaluate the shared weight self-attention, we first pre-train the BERT model using shared weight self-attention. Subsequently, we assess the pre-trained BERT model across a range of NLP tasks, including the General Language Understanding Evaluation (GLUE) Benchmark \cite{wang2018glue} and question-answering datasets such as SQuAD v1.1 \cite{rajpurkar2016squad} and SQuAD v1.2 \cite{rajpurkar2018know}. For our baseline comparison, we use the standard self-attention-based BERT model \cite{devlin2018BERT}, as well as the symmetric and pairwise-based self-attention in BERT models from \citet{courtois2024symmetric}.

\subsection{Pre-training Shared Attention Based BERT}
\textbf{Dataset: }To pre-train the shared weight attention-based BERT model, we utilized the same corpora as the standard BERT-base-uncased model, specifically the BooksCorpus (800M words) \cite{Zhu_2015_ICCV} and English Wikipedia (2,500M words), resulting in a total of approximately 3.2 billion tokens. 

\textbf{Pre-training Setup: }We adopt the configuration settings of the standard BERT model~\cite{devlin2018BERT}, which includes 12 layers, 768 hidden dimensions, and 12 attention heads. The maximum sequence length is set at 512 tokens. Regarding hyperparameters, we maintained the hidden dropout and attention dropout rates at 0.1. The pre-training is conducted over 20 epochs. 

We employ four H100 GPUs for computational resources, configuring each with a batch size of 132. The Adam optimizer \cite{kingma2014adam} was used, incorporating weight decay with $\beta_1 = 0.9$ and $\beta_2 = 0.999$. Masked language modeling is performed using a mask ratio of 0.15.

\begin{figure}[!t]

 \begin{center}
    \includegraphics[width=1.00\linewidth]{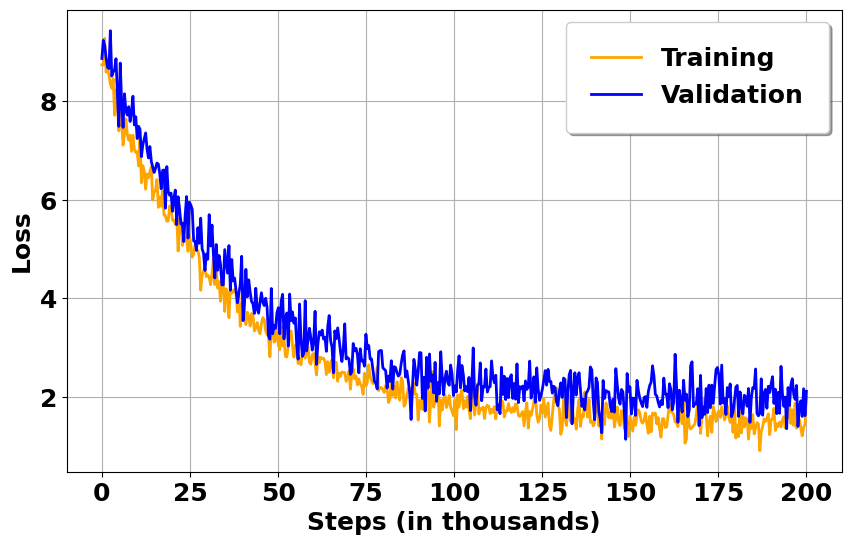}

   \end{center}
\caption{Pretraining loss curves for the shared weight self-attention mechanism. The plot shows the loss for both training and validation sets over 200,000 steps.}\label{fig:loss}
\end{figure}

\textbf{Pre-training Results}
Figure \ref{fig:loss} presents the training and validation loss curves during the pre-training of our shared self-attention based BERT model. Initially, the training and validation losses were high, starting at approximately $7.0$. This initial high loss is typical of models learning to adjust weights from random initialization.  As training progresses, the loss demonstrates a steady decline. After approximately $200,000$ steps, both the training and validation losses are significantly reduced, stabilizing at around $1.9$. 

\begin{table*}[ht]
\centering
\resizebox{0.90\textwidth}{!}{ 
\begin{tabular}{l|c|c|c|c|c|c|c|c|c}

\hline
\textbf{Model} & \textbf{MRPC} & \textbf{CoLA} & \textbf{MNLI (m/mm)} & \textbf{QQP} & \textbf{RTE} & \textbf{STSB} & \textbf{QNLI} & \textbf{SST-2} & \textbf{Average} \\
\hline

Standard&  87.27 & 52.64 & 81.66/82.07 & 88.86 & 59.42 & 88.19 & 88.76 & 90.92 & 79.97 \\
Symmetric &   78.36 & 49.22 & 78.66/79.05 & 87.70 & 53.43 & 84.47 & 86.90 & 89.56 & 76.37 \\
Pairwise & 87.83 & 51.91 & 81.60/82.02 & 88.89 & 59.58 & 86.88 & 88.78 & 89.78 & 79.03 \\ \hline
Shared Weight &  88.14 & 53.91 & 80.94/81.82& 88.24 & {59.60} & {88.78} & {88.02} & {89.84} & {79.92}\\
\hline
 \end{tabular}}
\caption{Performance comparison of different models across various GLUE benchmark tasks. The bold values indicate the best performance for each task. The evaluation metrics are accuracy for MRPC, MNLI, QQP, RTE, QNLI, and SST-2; Matthews correlation for CoLA; and Pearson/Spearman correlation for STS-B.}\label{tab:glue_results}
\end{table*}

\begin{table*}[ht]
\centering
\resizebox{0.70\textwidth}{!}{ 
\begin{tabular}{l|cc|cc|cc}
\hline
\multicolumn{1}{c|}{\textbf{Dataset}} & \multicolumn{2}{c|}{\textbf{SQuAD v1.1}} & \multicolumn{2}{c|}{\textbf{SQuAD v1.2}} & \multicolumn{2}{c}{\textbf{Average}} \\
\hline
\textbf{Model} & \textbf{EM} & \textbf{F1} & \textbf{EM} & \textbf{F1} & \textbf{EM} & \textbf{F1} \\
\hline
Standard (single) & 82.18 & 90.01 & 79.35 & 83.65 & 80.10 & 81.47 \\
Standard (single + TriviaQA) & 83.46 & 92.43 & 81.06 & {86.79} & 82.26 & 89.61 \\
\hline
Shared Weight (single) & 81.53 & 89.50 & 78.87 & 83.10 & 80.20 & 86.30 \\
Shared Weight (single + TriviaQA) & {83.19} & {91.97} & {80.16} & 85.78 &  {81.68} &  {88.88} \\
\hline
\end{tabular}} 
\caption{Comparison of EM and F1 scores on SQuAD v1.1 and v1.2}
\label{tab:results}
\end{table*}

\subsection{GLUE Benchmark}

 We evaluate our model on the GLUE Benchmarks \cite{wang2019glue} (Dataset description and hyperparameters in the Appendix \ref{data:description} and \ref{hp}).  
 
Table \ref{tab:glue_results} provides a comparison of the performance of various models, including standard, symmetric, pairwise, and shared, in the GLUE benchmark tasks.
We observe that the shared model consistently demonstrates superior or competitive performance compared to the other models across multiple tasks. Specifically, the shared model achieves approximately 0.87\% higher accuracy than the standard self-attention model for MRPC, about 9.78\% better performance than the symmetric self-attention model for CoLA, and approximately 2.0\% improvement over the pairwise self-attention model for the STS-B data set. Overall, the shared weight self-attention model exhibits improvements of -0.05\% +3.55\%, and +0.89\% over the standard, symmetric, and pairwise models, respectively, in terms of accuracy.

\subsection{Question Answering}

We utilize the SQuAD v1.1 \cite{rajpurkar2016squad} and SQuAD v2.0 \cite{rajpurkar2018know} datasets to evaluate the performance of our shared weight attention in the BERT model in answering questions.(Dataset description and hyperparameters in the Appendix \ref{data:description} and \ref{hp})

Table \ref{tab:results} shows the performance comparisons on the question-answering datasets. For the SQuAD v1.1 dataset, employing shared weight self-attention results in a decrease of 0.65\% in EM and 0.51\% in F1 score compared to the standard self-attention. However, when fine-tuning on the TriviaQA dataset \cite{joshi2017triviaqa}, we observe slight decreases of 0.27\% in EM and 0.46\% in F1 score. 

For the SQuAD v1.2 dataset, the use of shared self-attention results in a decrease of 0.48\% in EM and 0.52\% in F1 score compared to the standard self-attention. However, fine-tuning with the TriviaQA dataset leads to a decrease of 0.9\% in EM and 1.01\% in F1 score.

\begin{table*}[ht]
\centering
\resizebox{0.64\textwidth}{!}{ 
\begin{tabular}{c|c|c}
\hline
\textbf{Function} & \textbf{Expression} & \textbf{Parameters} \\  \hline
Standard & $\mathbf{Q}(X)\mathbf{K}(X)^T \cdot \mathbf{V}(X)$ & $3d^2$ \\ 
Symmetric & $\mathbf{Q}(X)\mathbf{Q}(X)^T \cdot \mathbf{V}(X)$ & $2d^2$ \\ 
Pairwise & $\mathbf{Q}(X)U\mathbf{Q}(X)^T \cdot \mathbf{V}(X)$ & $2d^2 + \frac{d^2}{m}$ \\ \hline
Shared Weight & $(\mathbf{S}(X) D_q)(\mathbf{S}(X) D_k)^T  \cdot (\mathbf{S}(X) D_v)$ & $d^2 + 3d$ \\ \hline
\end{tabular}}
\caption{Comparison of parameter counts in different attention mechanisms. Here $U$ is a matrix of pairwise factors, $m$ is the number of heads in the Transformer block.}
\label{tab:param}
\end{table*}

\begin{table}[ht]
\centering
\resizebox{0.49\textwidth}{!}{ 
\begin{tabular}{c|c|c}
\hline
\textbf{Config} & \textbf{Operator} & \textbf{Parameters} \\ \hline
\textbf{BERT\textsubscript{base}} & Standard & 109,514,298 \\ 
 & Symmetric & 102,427,194 (6.47\%) \\ 
 & Pairwise & 103,017,018 (5.93\%) \\ \cline{2-3} 
 & Shared Weight & 95,337,218 (12.94\%) \\ \hline
\end{tabular}
}
\caption{Parameter comparison for BERT configurations.}
\label{tab:BERT_config}
\end{table}

\section{Ablation Study}

\textbf{Parameter Analysis:} This study explores the efficiency of using shared weights in the self-attention mechanism. By implementing a shared transformation, \( \mathbf{S}(X) \), along with separate diagonal matrices \( D_q \), \( D_k \), and \( D_v \) for queries, keys, and values, the model requires fewer parameters, totaling \( \mathcal (d^2 + 3d) \). This setup results in a 66.53\% reduction in parameters compared to the traditional \( \mathcal(3d^2) \) needed by the standard self-attention in BERT, as highlighted in Table \ref{tab:param}. Integrating this approach into the overall BERT\textsubscript{base} model reduces the total number of parameters by 12.94\%, detailed in Table \ref{tab:BERT_config}. This significant decrease in parameters enhances the model’s computational efficiency without greatly affecting performance.

\begin{table*}
    \centering
    \resizebox{0.75\textwidth}{!}{ 
    \begin{tabular}{c|cc|cc|cc}
    \hline
\multicolumn{1}{c|}{\textbf{Dataset}} & \multicolumn{2}{c|}{\textbf{MNLI}} & \multicolumn{2}{c|}{\textbf{QQP}} & \multicolumn{2}{c}{\textbf{SST-2}} \\
\hline
\textbf{Noise} & \textbf{Standard} & \textbf{Shared Weight} & \textbf{Standard} & \textbf{Shared Weight} & \textbf{Standard} & \textbf{Shared Weight} \\
\hline

        0\% & 81.66 & 80.94 & 88.86 & 88.24 & 90.92 & 89.84 \\
        5\% & 80.02 & 80.85 & 88.10 & 88.14 & 89.34 & 89.03 \\
        10\% & 79.42 & 80.02 & 85.63 & 87.43 & 91.03 & 90.34 \\
        15\% & 78.53 & 80.11 & 84.23 & 87.00 & 88.56 & 89.43 \\
        20\% & 77.42 & 79.82 & 83.98 & 87.16 & 86.34 & 88.18 \\
        25\% & 74.53 & 78.42 & 81.32 & 85.32 & 85.14 & 88.81 \\
        30\% & 72.47 & 77.12 & 80.72 & 85.52 & 83.52 & 84.35 \\
        35\% & 70.34 & 76.94 & 77.70 & 84.63 & 80.48 & 83.19 \\
        40\% & 68.24 & 75.19 & 75.24 & 82.54 & 74.35 & 82.52 \\
        \hline
    \end{tabular}
    }
    \caption{Performance comparison of traditional and shared weight self-attention models under various noise levels on MNLI, QQP, and SST-2 datasets.}\label{tab:robustness}
\end{table*}

\textbf{Robustness Analysis:} We test the robustness of our shared weight self-attention mechanism against traditional self-attention using the GLUE benchmark datasets (MNLI, QQP, SST-2). To simulate noise, we compute the average $L_2$ norm of the input embeddings and introduce spherical Gaussian noise with a standard deviation of 1, which corresponds to approximately 0\% to 40\% of the input embedding norm.  The performance is summarized in Table \ref{tab:robustness}. The results show that the shared weight self-attention model maintains higher accuracy under noisy conditions. For instance, on the MNLI dataset, while the accuracy of the standard model drops from 81.66\% to 68.24\% with increasing noise, the shared model decreases less sharply, from 80.94\% to 75.19\%. This pattern of greater resilience is consistent across other datasets like QQP and SST-2.

\begin{figure}[!t]

 \begin{center}
    \includegraphics[width=1.00\linewidth]{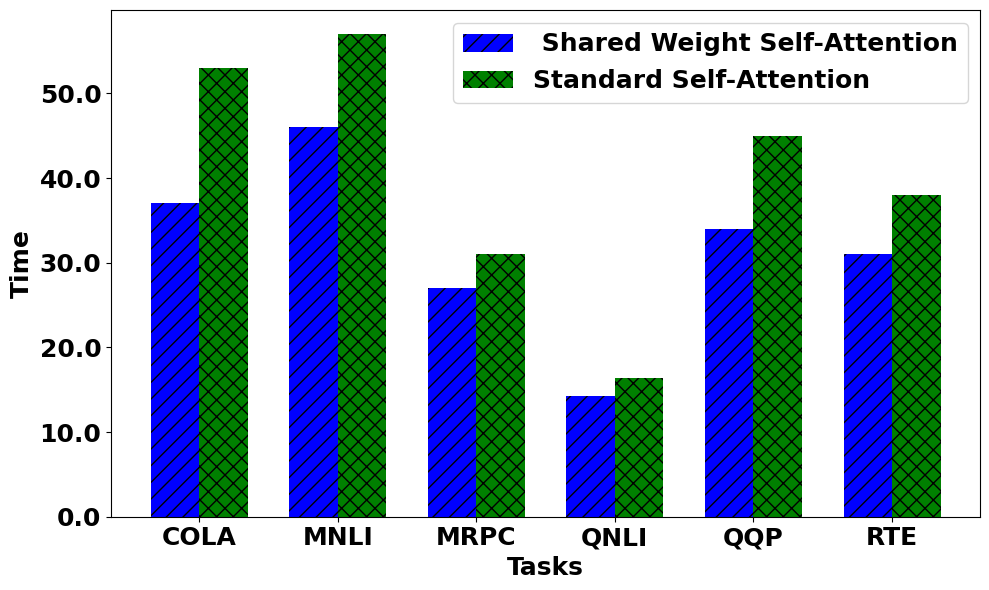}

   \end{center}
\caption{Training Time Comparison Between shared Weight and standard self-attention on GLUE tasks. CoLA, MRPC, and QQP are recorded in seconds, and Other tasks are presented in minutes.}\label{fig:Training_Time}
\end{figure}

\textbf{Training Time:} We assess the efficiency of shared weight self-attention compared to traditional self-attention mechanisms across six NLP tasks: CoLA, MNLI, MRPC, QNLI, RTE, and QQP in Figure~\ref{fig:Training_Time}. Our findings indicate substantial improvements in processing times for each task. For instance, in the CoLA task,  shared weight self-attention reduced processing time by 30\%, from 53 to 37 seconds, increasing speed by approximately 43\%. Similar enhancements are seen in other tasks: MNLI’s time was reduced by 19\%, MRPC by 12\%, QNLI by 11\%, RTE by 18\%, and QQP by 13\%.

Each task is executed for one epoch with a batch size of 16, highlighting the efficiency gains from shared weight self-attention. These improvements suggest the potential for significant cost savings and enhanced productivity. Tests were performed using an NVIDIA RTX A6000 GPU with 50GB of VRAM.

\begin{table*}[ht]
\centering
\resizebox{0.94\textwidth}{!}{ 
\begin{tabular}{c|cc|cc|cc|cc}
\hline
\multicolumn{1}{c|}{\textbf{Domain}} & \multicolumn{2}{c|}{\textbf{MNLI}} & \multicolumn{2}{c|}{\textbf{QNLI}} & \multicolumn{2}{c|}{\textbf{QQP}} & \multicolumn{2}{c}{\textbf{MRPC}} \\ \hline
          & \textbf{Standard} & \textbf{Shared Weight}    & \textbf{Standard} & \textbf{Shared Weight}     & \textbf{Standard} & \textbf{Shared Weight}    & \textbf{Standard} & \textbf{Shared Weight}     \\ \hline
\textbf{MNLI} & 81.66  & 80.94  & 72.24  & 74.30   & 49.03  & 50.21  & 60.12  & 69.03  \\ \hline
\textbf{QNL}  & 77.99  & 78.7   & 86.76  & 89.02   & 72.31  & 51.62  & 53.87  & 50.29  \\ \hline
\textbf{QQP}  & 59.21  & 58.42  & 52.71  & 54.92   & 88.86  & 88.89  & 62.03  & 67.21  \\ \hline
\textbf{MRPC} & 62.83  & 62.88  & 59.21  & 52.32   & 68.30  & 78.76  & 82.27  & 88.14  \\ \hline
\end{tabular}}
\caption{Comparison of model performance on MNLI, QNLI, QQP, and MRPC tasks under standard and shared weight conditions, highlighting cross-task adaptability.}

\label{table:domain_adaptation}
\end{table*}

\textbf{Cross-Domain Performance:}
Table~\ref{table:domain_adaptation} illustrates the performance of NLP models under two conditions: standard and shared weight, across four different tasks—MNLI, QNLI, QQP, and MRPC. The highest performance is typically observed within the same domain (diagonal entries), demonstrating that models are most effective on the data they are trained on. The shared weight condition generally enhances cross-domain performance, indicating the utility of parameter sharing for generalization across related tasks. For instance, MNLI trained models show improved performance on QNLI and MRPC in the shared Weight scenario.

\textbf{Comparison of Self-Attention Mechanisms}
\label{sec:comparison}

Table~\ref{tab:attention_comparison} presents a comparative analysis of various self-attention mechanisms, including standard, symmetric, pairwise, partial QK sharing, and the proposed full QKV sharing. Standard self-attention employs three separate weight matrices, \(W_q\), \(W_k\), and \(W_v\), resulting in the highest parameter count (\(3d^2\)) and computational complexity. Symmetric and partial QK sharing reduce parameters by sharing query and key matrices, achieving a 33\% reduction but compromising expressiveness. Pairwise attention enhances token interactions with an additional matrix \(U\), increasing complexity while providing moderate efficiency gains. In contrast, full QKV sharing employs a single weight matrix \(W_s\) with diagonal scaling matrices \(D_q\), \(D_k\), and \(D_v\), reducing parameters by 66.67\%, lowering computational overhead, and retaining expressiveness. This approach also improves training speed by 15-20\%, enhances noise robustness, and simplifies implementation, making it a more efficient and effective alternative to other self-attention variants.

\begin{table*}[ht!]
\centering

\scalebox{.70}{
\begin{tabular}{l|c|c|c|c|c}
\hline
\textbf{Feature}              & \textbf{Standard} & \textbf{Symmetric} & \textbf{Pairwise} & \textbf{Partial QK Sharing} & \textbf{Full QKV Sharing} \\
\hline
\textbf{Weight Matrices}      & $W_q, W_k, W_v$  & $W_q = W_k, W_v$  & $W_q, U, W_v$   & $W_q = W_k, W_v$          & Single $W_s$              \\
\hline
\textbf{Parameter Count}      & $3d^2$           & $2d^2$             & $2d^2 + {d^2}/m$       & $2d^2 + d$                & $d^2 + 3d$               \\
\hline
\textbf{Parameter Reduction}  & 0\%              & 33\%               & 30-35\%          & 33\%                      & 66.67\%                  \\
\hline
\textbf{Computational Complexity} & High        & Moderate           & High             & Moderate                  & Low                      \\
\hline
\textbf{Diagonal Scaling Matrices} & No         & No                 & No               & No                        & Yes                      \\
\hline
\textbf{Expressiveness}       & High             & Reduced Q-K diversity & Enhanced (pairwise) & Moderate               & Retained (via scaling)   \\
\hline
\textbf{Training Speed}       & Baseline         & 10-15\% faster     & 5-10\% slower    & 10-15\% faster            & 15-20\% faster           \\
\hline
\textbf{Memory Usage}         & High             & Moderate           & High             & Moderate                  & Low                      \\
\hline
\textbf{Implementation Simplicity} & Complex     & Simple             & Complex (U matrix) & Simple                  & Simplest                 \\
\hline
\end{tabular}
}
\caption{Comparison of Different Self-Attention Methods} \label{tab:attention_comparison}
\end{table*}

\section{Related Work}

The introduction of self-attention in Transformer architecture in 2017 by \citet{vaswani2017attention} marked a significant turning point by enabling models to dynamically concentrate on relevant parts of input sequences, building upon earlier work by \citet{bahdanau2014neural}, who applied attention mechanisms within recurrent neural networks (RNNs) for machine translation and thus improved translation accuracy.

According to \citet{luong2015effective}, self-attention mechanisms were enhanced to better model complex data dependencies, which contributed to the development of more advanced attention models. Subsequently, \citet{vaswani2017attention} delved deeper into self-attention mechanisms, resulting in the creation of models like BERT \cite{devlin2018BERT}. This model utilized bidirectional training of Transformers to capture context from both directions in a sequence, leading to state-of-the-art performance across a range of NLP tasks.

Reviews conducted by \citet{galassi2020attention} and \citet{niu2021review} highlighted the significant role of weighted dot-product attention in contemporary models. \citet{guo2022attention} assessed the versatility of self-attention mechanisms in computer vision, demonstrating their utility beyond NLP. To enhance the efficiency of attention mechanisms, \citet{child2019generating} presented the sparse Transformer, which reduces the complexity of full attention mechanisms for more efficient long-sequence processing. \citet{beltagy2020longformer} introduced the Longformer, which utilizes dilated sliding window attention to efficiently handle longer context sequences.

\citet{he2023simplifying} presented a streamlined Transformer architecture that reduced model weight by 15\% without compromising performance. In a subsequent study, \citet{courtois2024symmetric} introduced a pairwise compatibility operator that enhanced the dot-product method with a shared linear operator and a bilinear matrix, thereby improving token interactions and BERT model performance.

Our proposed method builds upon these advancements by utilizing a single shared weight matrix, $\mathbf{W}_s$, for a unified representation. Keys, Queries, and Values are derived through diagonal matrix multiplication with specific vectors, resulting in a 66.53\% reduction in parameters within the self-attention block. Despite this significant reduction, our method maintains robust performance across BERT configurations, demonstrating the potential for more efficient yet powerful NLP models.

\section{Limitations}

Our work mainly focused on studying an alternative compatibility function with the self-attention mechanism in transformer-based encoder models, particularly those evaluated using NLU. While we show good performance in this setting, our results do not necessarily translate to decoder models, pure language modeling tasks, or machine translation. For many applications, the cross-attention mechanism is crucial for achieving high accuracy on these tasks and does not completely align with our use case, where we support shared representations through a trainable matrix. In our model, we use a single shared weight matrix $\mathbf{W}_s$ for the unified representation, reducing the number of parameters in the self-attention block by 66. 67\% compared to the baseline models. Although this reduction is significant, its impact on broader applications requires further analysis. Due to the resource restriction, we only observed improved training efficiency for smaller BERT-like models with approximately 100 million parameters in one of our experiments. However, these findings may not generalize well to much larger models, such as those of an order of magnitude larger. One limitation of our approach is its reliance on a single softmax weight, which may not exhibit optimal behavior for more complex datasets, suggesting the need for multiple weights or alternative strategies. We also recognize the importance of decoder components in text-generation tasks, which we have yet to fully explore. Overcoming these challenges through future investigations will contribute to the generalization and scalability of our approach in diverse NLP frameworks.

\section{Conclusions}
The shared weight self-attention mechanism presented simplifies the traditional self-attention model by using a single shared matrix with element-wise scaling for keys, queries, and values. This approach reduces parameter complexity while maintaining high performance. Extensive experiments on the GLUE benchmark datasets demonstrate that the shared weight self-attention-based Bert model performs comparably to traditional Bert models on clean data and shows superior robustness under various noise conditions. The empirical results highlight the model's ability to capture essential features more effectively and maintain stability even with noisy inputs. This makes the shared weight self-attention mechanism particularly suitable for applications in environments with noisy or imperfect data. Additionally, the significant reduction in learnable parameters leads to more efficient models that are easier to deploy in resource-constrained settings.

\bibliography{custom} 

\appendix

\section{ Appendix}
\label{sec:appendix}
\subsection{Limitations}

Our work mainly focused on studying an alternative compatibility function with the self-attention mechanism in transformer-based encoder models, particularly those evaluated using NLU. While we show good performance in this setting, our results do not necessarily translate to decoder models, pure language modeling tasks, or machine translation. For many applications, the cross-attention mechanism is crucial for achieving high accuracy on these tasks and does not completely align with our use case, where we support shared representations through a trainable matrix. In our model, we use a single shared weight matrix $\mathbf{W}_s$ for the unified representation, reducing the number of parameters in the self-attention block by 66. 67\% compared to the baseline models. Although this reduction is significant, its impact on broader applications requires further analysis. We observed improved training efficiency for smaller BERT-like models with approximately 100 million parameters in one of our experiments. However, these findings may not generalize well to much larger models, such as those of an order of magnitude larger. Our models were benchmarked with GLUE and the newer SuperGLUE, providing better evaluation metrics for current models. One limitation of our approach is its reliance on a single softmax weight, which may not exhibit optimal behavior for more complex datasets, suggesting the need for multiple weights or alternative strategies. We also recognize the importance of decoder components in text-generation tasks, which we have yet to fully explore. Overcoming these challenges through future investigations will contribute to the generalization and scalability of our approach in diverse NLP frameworks.

\subsection{Dataset Description} \label{data:description}

We evaluate our shared weight self-attention mechanism on multiple tasks from the GLUE benchmark \cite{wang2018glue}. Specifically, our method is tested on the following datasets: CoLA, SST-2, MRPC, STS-B, QQP, MNLI, QNLI, and RTE. To assess the question-answering capabilities of our approach, we use the SQuAD v1.1 \cite{rajpurkar2016squad} and SQuAD v2.0 \cite{rajpurkar2018know} datasets. These datasets consist of question-answer pairs derived from Wikipedia articles, providing a robust basis for evaluating the performance of question-answering models. The datasets used in this study are listed in Table \ref{tab:dataset_statistics}.

\begin{table}[ht]
    \centering
    \begin{tabular}{c|c|c|c}
        \hline
        Dataset & \# Train & \# Validation & \# Test \\ 
        \hline
        SQuAD v1.1 & 87.6k & 10.6k & - \\
        \hline
        SQuAD v2.0 & 130k & 11.9k & - \\
        \hline
        CoLA & 8.55k & 1.04k & 1.06k \\
        \hline
        SST2 & 67.3k & 872 & 1.82k \\
        \hline
        MRPC & 3.67k & 408 & 1.73k \\
        \hline
        STS-B & 5.75k & 1.5k & 1.38k \\
        \hline
        QQP & 364k & 40.4k & 391k \\
        \hline
        MNLI & 393k & 9.8k & 9.8k \\
        \hline
        QNLI & 105k & 5.46k & 5.46k \\
        \hline
        RTE & 2.49k & 277 & 3k \\
        \hline
    \end{tabular}
    \caption{Dataset Statistics}
    \label{tab:dataset_statistics}
\end{table}

\subsection{Evaluation Metric}
We employ the Matthews correlation for CoLA, Pearson and Spearman correlation for STS-B, average matched accuracy and F1 score for MNLI, and accuracy for other NLU tasks.

\subsection{Hyperparameter} \label{hp}

For the GLUE benchmark, uniform hyperparameters are consistently implemented across all tasks to ensure comparability and consistent results. Specifically, the attention dropout and weight decay rates are uniformly set at 0.1, while the initial learning rate is fixed at \(1 \times 10^{-4}\). Subsequently, the learning rate is fine-tuned to \(2 \times 10^{-5}\) and \(2 \times 10^{-6}\). Each dataset is trained for 10 epochs to attain optimal performance.

For the SQuAD datasets, the dropout rate is fixed at 0.2, while the attention dropout rate is set at 0.05, and the weight decay rate is established at 0.1. The initial learning rate is set at $1 \times 10^{-4}$, which is subsequently adjusted to $2 \times 10^{-5}$ and $2 \times 10^{-6}$. Training is conducted over a period of 5 epochs.

\end{document}